\documentclass{article} 
\usepackage{nips}
\usepackage{times}
\usepackage[utf8]{inputenc} 
\usepackage[T1]{fontenc}    
\usepackage{hyperref}       
\usepackage{url}            
\usepackage{booktabs}       
\usepackage{nicefrac}       
\usepackage{microtype}      
\usepackage[page]{appendix}
\usepackage{amsmath,amsfonts,amssymb,hyperref,multirow}
\usepackage{graphicx}
\usepackage{xcolor}

\graphicspath{{}{figures/}}

\def\R{\mathbb{R}}
\def\E#1{\mathbb{E}\!\left[{#1}\right]}
\def\V#1{\mathrm{var}\!\left[{#1}\right]}
\def\M#1{\left<{#1}\right>}
\def\D#1#2{\frac{\partial #1}{\partial #2}}

\def\argmin{\mathop{\rm argmin}}
\def\Iota#1#2{\left\{{#1}\dots{#2}\right\}}
\def\ij{{ij}}
\def\zj{{0j}}
\def\v{\mathbf{v}}
\def\w{\mathbf{w}}

\def\ip{{i^\prime}}
\def\jp{{j^\prime}}
\def\ipjp{{i^\prime j^\prime}}

\def\of#1{^{#1}}

\makeatletter
\renewcommand{\section}{%
  \@startsection{section}{1}{\z@}%
                {-2.0ex \@plus -0.5ex \@minus -0.2ex}%
                { 1.0ex \@plus  0.2ex \@minus  0.2ex}%
                {\large\bf\raggedright}%
}
\renewcommand{\subsection}{%
  \@startsection{subsection}{2}{\z@}%
                {-1.2ex \@plus -0.5ex \@minus -0.2ex}%
                { 0.6ex \@plus  0.2ex}%
                {\normalsize\bf\raggedright}%
}
\makeatother

\title{Diagonal Rescaling For Neural Networks}
\author{Jean Lafond, Nicolas Vasilache, L\'eon Bottou \\
  Facebook AI Research, New York \\
  \texttt{lafond.jean@gmail.com, ntv@fb.com, leon@bottou.org}}

\begin{document}
\maketitle

\begin{abstract}
  We define a second-order neural network stochastic gradient training
  algorithm whose block-diagonal structure effectively amounts to
  normalizing the unit activations. Investigating why this algorithm
  lacks in robustness then reveals two interesting insights. The first
  insight suggests a new way to scale the stepsizes, clarifying
  popular algorithms such as RMSProp as well as old neural network
  tricks such as fanin stepsize scaling. The second insight stresses
  the practical importance of dealing with fast changes of the
  curvature of the cost.
\end{abstract}

\section{Introduction}

Although training deep neural networks is crucial for their
performance, essential questions remain unanswered. 
Almost everyone nowadays trains convolutional neural networks (CNNs)
using a canonical bag of tricks such as dropouts, rectified linear
units (ReLUs), and batch normalization
\citep{dahl-2013,ioffe-szegedy-2015}. Accumulated empirical evidence
unambiguously shows that removing one of these tricks
leads to less effective training.

Countless papers propose new additions to the canon. Following the
intellectual framework set by more established papers, the proposed
algorithmic improvements are supported by intuitive arguments and
comparative training experiments on known tasks.  This approach is
problematic for two reasons.  First, the predictive value of intuitive
theories is hard to assess when they share so little with each
other. Second, the experimental evidence often conflates two important
but distinct questions: which learning algorithm works best when
optimally tuned, and which one is easier to tune.

We initially hoped to help the experimental aspects by offering a
solid baseline in the form of an efficient and well understood way to
tune a simple stochastic gradient (SG) algorithm, hopefully with a
performance that matches the canonical bag of tricks. To that effect,
we consider reparametrizations of feedforward neural networks that are
closely connected to the normalization of neural network activations
\citep{schraudolph-2012,ioffe-szegedy-2015} and are amenable to zero
overhead stochastic gradient implementations.  Invoking the usual
second order optimization arguments
\citep{becker-lecun-1989,ollivier-2013,desjardins-2015,caron-ollivier-2016}
leads to tuning the reparametrization with a simple diagonal or
block-diagonal approximation of the inverse curvature matrix. The
resulting algorithm performs well enough to produce appealing training
curves and compete favorably with the best known methods. However this
algorithm lacks robustness and occasionally diverges with little
warning. The only way to achieve robust convergence seems to reduce
the global learning rate to a point that negates its speed benefits.

Our critical investigation led to the two insights that constitute the
main contributions of this paper.  The first insight provides an
elegant explanation for popular algorithms such as RMSProp
\citep{tieleman-hinton-2012} and also clarifies well-known stepsize
adjustments that were popular for the neural networks of the
1990s. The second insight explains some surprising aspects of batch
normalization \cite{ioffe-szegedy-2015}. These two insights provide a
unified perspective in which we can better understand and compare how
popular deep learning optimization techniques achieve efficiency
gains.

This document is organized as follows.  Section~\ref{sec:zerooverhead}
describes our reparametrization scheme for feedforward neural networks
and discusses the efficient implementation of a SG algorithm.
Section~\ref{sec:stepsizes} revisits the notion of stepsizes when one
approximates the curvature by a diagonal or block-diagonal matrix.
Section~\ref{sec:fastcurvature} shows how fast curvature changes can
derail many second order optimization methods and justify why it is
attractive to evaluate curvature on the current minibatch as in
batch-normalization.

\section{Zero overhead reparametrization}
\label{sec:zerooverhead}

This section presents our reparametrization setup for the trainable
layers of a multilayer neural network.  Consider a linear
layer\footnote{\relax Appendix~\ref{app:convol} discusses the case of
  convolutional layers.}  with $n$ inputs $x_i$ and $m$ outputs $y_j$
\begin{equation}
  \label{eq.linear}
    \forall j\in\Iota1m \qquad y_j = w_\zj + \sum_{i=1}^n x_i w_\ij~.
\end{equation}
Let $E$ represent the value of the loss function for the current example.
Using the notation $g_j = \D{E}{y_j}$ and the convention $x_0=1$, we can write
\[
\forall (i,j)\in\Iota0n\times\Iota1m \qquad \D{E}{w_\ij} = x_i g_j~.
\]

\subsection{Reparametrization}

We consider reparametrizations of \eqref{eq.linear} of the form
\begin{equation}
  \label{eq:reparam}
  y_j = \beta_j \big( v_\zj + \sum_{i=1}^n \alpha_i (x_i-\mu_i) v_\ij \big) ~,\\
\end{equation}
where $v_\zj$ and $v_\ij$ are the new parameters and $\mu_i$, $\alpha_i>0$,
and $\beta_j>0$ are constants that specify the exact reparametrization.
The old parameters can then be derived from the new parameters with the relations
\[
\begin{aligned}
  w_\ij &= \alpha_i\beta_j v_\ij \qquad \text{(~for $i=1\dots n$.~)}\\
  w_\zj &= \beta_j v_\zj - \sum_{i=1}^{n} \mu_i w_\ij = \beta_j v_\zj - \sum_{i=1}^{n} \mu_i \alpha_i \beta_j v_\ij ~.
\end{aligned}
\]
Using the convention $z_0=1$ and $z_i=\alpha_i(x_i-\mu_i)$, we can compactly write
\[
   y_j = \beta_j \sum_{i=0}^n v_\ij z_i ~.
\]
Running the SG algorithm on the new parameters amounts
to updating these parameters by adding a quantity proportional to
\[ 
\delta v_\ij = \M{\D{E}{v_\ij}} = \M{\beta_j g_j z_i} ~,
\]
where the notation $\M{\dots}$ is used to represent
an averaging operation over a batch of examples.
The corresponding modification of the old parameters is
then proportional to

\begin{equation}
  \label{eq:quasidiagonal}
  \begin{aligned}
  \delta w_\ij &= \alpha_i \beta_j \delta v_\ij
     ~=~  \M{\beta_j^2 g_j \: \alpha_i^2 (x_i - \mu_i) } \\
  \delta w_\zj &= \beta_j \delta v_\zj - \sum_{i=0}^{n} \mu_i \delta w_\ij 
     ~=~ \left< \beta_j^2 g_j \right> - \sum_{i=1}^{n} \mu_i \delta w_\ij~.
  \end{aligned}
\end{equation}

This means that we do not need to store the new parameters.  We can
perform both the forward and backward computations using the usual
$w_\ij$ parameters, and use the above equations during the weight
update. This approach ensures that we can easily change the constant
$\alpha_i$, $\mu_i$, and $\beta_j$ at any time without changing the
function computed by the network.

Updating the weights using \eqref{eq:quasidiagonal} is very cheap
because we can precompute \mbox{$\beta_j^2g_j$} and
\mbox{$\alpha_i^2(x_i-\mu_i)$} in time proportional to $n+m$. This
overhead is negligible in comparison to the remaining computation
which is proportional to~$nm$.

\subsection{Block-diagonal representation}

The weight updates $\delta w_\ij$ described by
equation \eqref{eq:quasidiagonal} can also be 
obtained by pre-multiplying the averaged gradient
vector $\M{\partial E/\partial w_\ij}=\M{g_jx_i}$
by a specific block diagonal positive symmetric matrix.
Each block of this pre-multiplication reads as
\begin{equation*}
  \left[ \begin{array}{c}
      \delta w_\zj \\
      \delta w_{1j} \\
      \vdots \\
      \delta w_{nj}
    \end{array} \right]
  ~=~
  \beta_j^2 \left[ \begin{array}{cccc}
      1+\sum \alpha_i^2\mu_i \rule[-2ex]{0pt}{1ex}
      & -\alpha_1^2\mu_1 & \hdots & -\alpha_n^2\mu_n \\
      -\alpha_1^2\mu_1 & \alpha_1^2 &  &  \\
      \vdots & & \ddots & \text{\smash{\raisebox{2ex}{\large0}}} \\
      -\alpha_n^2\mu_n & \text{\smash{\large0}} & & \alpha_n^2
    \end{array} \right ] ~\times~
  \left[ \begin{array}{c}
      \M{ \partial E / \partial w_\zj } \\
      \M{ \partial E / \partial w_{1j} } \\
      \vdots \\
      \M{  \partial E / \partial w_{nj} } 
    \end{array} \right] .
\end{equation*}
This rewrite makes clear that the reparametrization \eqref{eq:reparam}
is an instance of \emph{quasi-diagonal} rescaling \citep{ollivier-2013},
with the additional constraint that, up to a scalar coefficient $\beta_j^2$,
all the blocks of the rescaling matrix are identical within a same layer.

\subsection{Choosing and adapting the reparametrization constants}
\label{sec:secondorder}

Many authors have proposed \emph{second order} stochastic gradient
algorithms for neural networks
\citep{becker-lecun-1989,park-amari-2000,
  ollivier-2013,martens-grosse-2015,caron-ollivier-2016}. Such
algorithms rescale the stochastic gradients using a suitably
constrained positive symmetric matrix. In all of these works, the key
step consists in defining an approximation $G$ of the curvature of the
cost function, such as the Hessian matrix or the Fisher Information
matrix, using ad-hoc assumptions that ensure that its inverse $G^{-1}$
is easy to compute and satisfies the desired constraints on the
rescaling matrix.

We can use the same strategy to derive sensible values for our
reparametrization constants.
Appendix \ref{app:curvature} derives a block-diagonal
approximation $G$ of the curvature of the cost function with respect
to the parameters $v_\ij$. Each diagonal block $G_j$ of this matrix
has coefficients
\begin{equation}
  \label{eq:canonicalng}
  \left[G_j\right]_{i\ip} ~=~ \beta_j^2 \E{g_j^2} \times \left\{ \begin{array}{ll}
    \E{z_i^2} \rule[-1.5ex]{0pt}{0pt} & \text{if $i=\ip$,} \\
    \E{z_i}\E{z_\ip} & \text {if $i\neq\ip$,}
    \end{array}\right.
\end{equation}
where the expectation $\E{\cdot}$ is meant with respect to
the distribution of the training examples.
Choosing reparametrization constants  $\mu_i$, $\alpha_i$, and $\beta_j$
that make this surrogate matrix equal to the identity amounts to ensuring
that a simple gradient step in the new parameters $v_\ij$ is
equivalent to a second order step in the original parameters $w_\ij$.
This is achieved by choosing
\begin{equation}
  \label{eq:canonicalconstants}
  \mu_i = \E{x_i} \qquad
  \alpha^2_i=\frac{1}{\V{x_i}} \qquad
  \beta^2_j=\frac{1}{\E{g_j^2}}~.
\end{equation}

It is not a priori obvious that we can continuously adapt the
reparametrization constants on the basis of the observed statistics
without creating potentially nefarious feedback loops in the
optimization dynamics. On the positive side, it is well-known that
pre-multiplying the stochastic gradients by a rescaling matrix
provides the usual convergence guarantees if the eigenvalues of the
rescaling matrix are upper and lower bounded by positive values
\citep[\S 4.1]{bottou-2016}, something easily achieved by adequately
restricting the range of values taken by the reparametrization
constants~$\alpha^2_i$, $\mu_i$, and~$\beta^2_j$. On the negative side,
since the purpose of this adaptation is to make sure the rescaling
matrix improves the convergence speed, we certainly do not want to see
reparametrization constants hit their bounds, or, worse, bounce
between their upper and lower bounds.

The usual workaround consists in ensuring that the
rescaling matrix changes very slowly. In the case of our reparametrization
scheme, after processing each batch of examples,
we simply update online estimates of the moments
\[
  \begin{array}{rclcl}
   \mathtt{mx[i]} &\leftarrow& \lambda\, \mathtt{mx[i]} &+& (1-\lambda)\,\M{x_i} \\[.3ex]
   \mathtt{mx2[i]} &\leftarrow& \lambda\, \mathtt{mx2[i]} &+& (1-\lambda)\,\M{x^2_i} \\[.3ex]
   \mathtt{mg2[j]} &\leftarrow& \lambda\, \mathtt{mg2[j]} &+& (1-\lambda)\,\M{g_j^2},
  \end{array}
\]
with $\lambda\approx0.95$, and we recompute the reparametrization
constants~\eqref{eq:canonicalconstants}. We additionally make sure
that their values remain in a suitable range. This procedure is
justified if we believe that the essential statistics of the $x_i$ and
$g_j$ variables change sufficiently slowly during the optimization.

\subsection{Informal comment about the algorithm performance}
\label{sec:notrobust}

This algorithm performs well enough to produce appealing training
curves and compete favorably with the best known methods (\emph{at least
for the duration of a technical paper}). The day-to-day practice
suggests a different story which is both more important and difficult
to summarize with experimental results. Finding a proper
stepsize with plain SG is relatively easy because excessive stepsizes
immediately cause a catastophic divergence. This is no longer the case
with this proposed algorithm: many stepsizes appear to work
efficiently, but occasionally cause divergence with little warning.
The only way to achieve robust convergence seems to be to reduce the
stepsize to a point that essentially negates the initial speed
gain. This observation does not seem to be specific to our particular
algorithm. For instance, \citet[][\S9.1]{lecun-etal-1998} mention
that, in practice, their diagonal rescaling method reduces the number
of iterations by no more than a factor of three relative to plain SG,
barely justifying the overhead.

\section{Stepsizes and diagonal rescaling}
\label{sec:stepsizes}

The difficulty of finding good global stepsizes with second order
optimization methods is in fact a well-known issue in optimization,
only made worse by the stochastic nature of the algorithms we
consider. After presenting a motivating example, we return to the
definition of the stepsizes and develop an alternative formulation
suitable for diagonal and block-diagonal rescaling approaches.

\subsection{Motivating example}
\label{sec:motiv}

\begin{figure}[t]
  \centering\vspace{-0.05\linewidth}
  \includegraphics[width=0.5\linewidth]{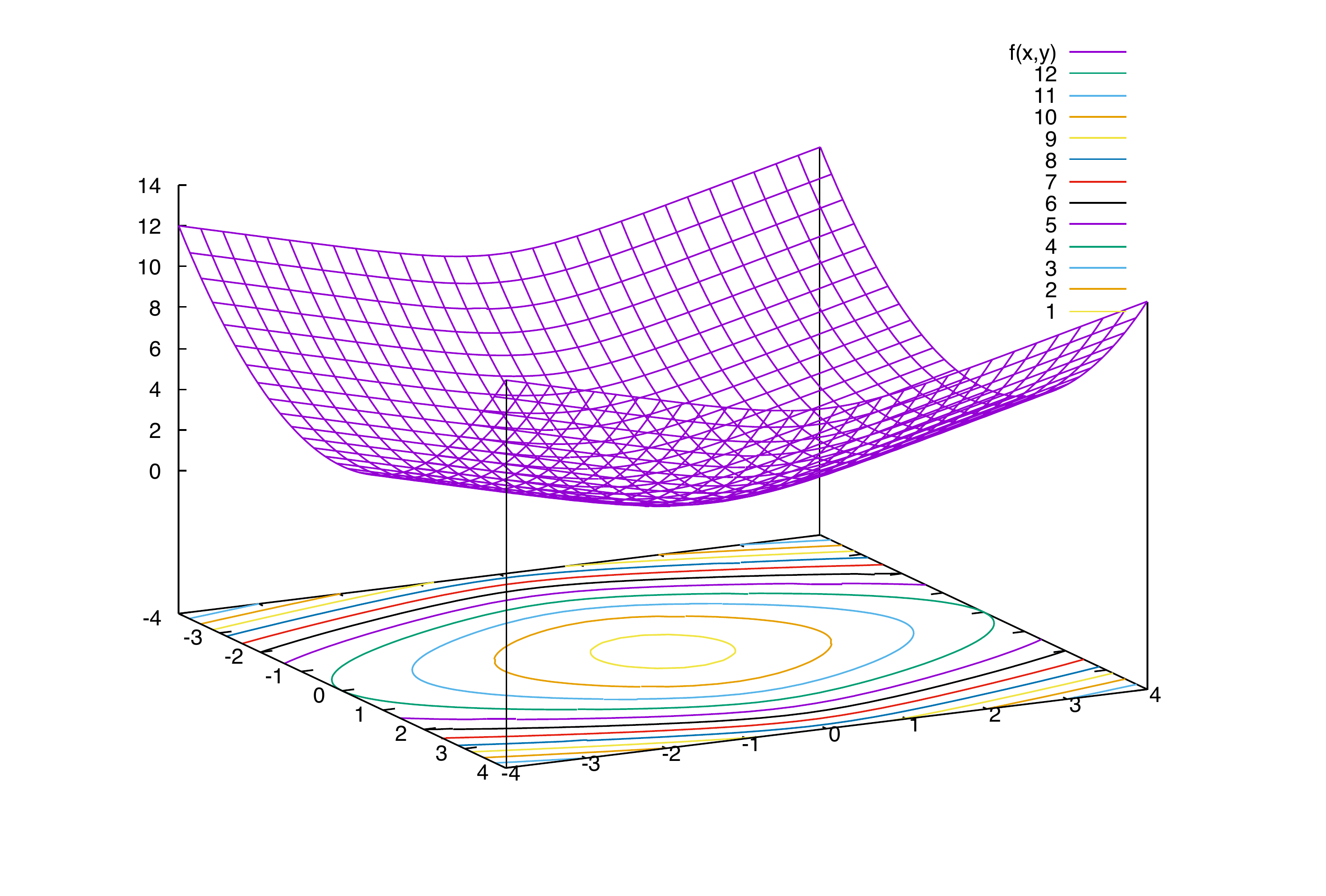}
  \par\vspace{-0.05\linewidth}
  \caption{The function $F(w_1,w_2)=\frac12 w_1^2 + \log(e^{w_2}+e^{-w_2})$.}
  \label{fig:convexf}
\end{figure}

Figure~\ref{fig:convexf} represents the apparently benign convex function
\begin{equation}
  \label{eq:benign}
  F(w_1,w_2) = \frac12 w_1^2 + \log(e^{w_2}+e^{-w_2})
\end{equation}
whose gradients and Hessian matrix respectively are
\[
\nabla F(w_1,w_2) = \left[ \begin{array}{c} w_1 \\ \mathrm{tanh}(w_2) \end{array} \right] \qquad
\nabla^2 F(w_1,w_2) = \left[ \begin{array}{cc} 1 & 0 \\ 0 & \mathrm{cosh}(w_2)^{-2} \end{array} \right]~.
\]
Following \citet[][\S6.5]{bottou-2016}, assume we are optimizing this
function with starting point $(3,3)$.  The first update moves the
current point along direction $-\nabla F\approx[-3,-1]$ which
unfortunately points slightly away from the optimum $(0,0)$. Rescaling
with the inverse Hessian yields a substantially worse direction
$-(\nabla^2F)^{-1}\nabla F\approx[-3,-101]$. The large second
coefficient calls for a small stepsize. Using stepsize
$\gamma\approx0.03$ moves the current point to $(2.9,0)$. Although the
new gradient $\nabla F\approx[-2.9,0]$ points directlty towards the
optimum, the small stepsize that was necessary for the previous update
is now ten times too small to effectively leverage this good situation.

We can draw two distinct lessons from this example:
\begin{enumerate}
\item[a)] A global stepsize must remain small enough to accomodate the
  most ill-conditioned curvature matrix met by the algorithm
  iterates. This is precisely why most batch second-order optimization
  techniques rely on line search techniques instead of fixing a single
  global stepsize \citep{nocedal-wright-2006}, something not easily
  done in the case of a stochastic algorithm. Therefore it is
  desirable to automatically adjust the stepsize to account for the
  conditioning of the curvature matrix.
\item[b)] The objective function \eqref{eq:benign} is a sum of terms
  operating on separate subsets of the variables. Absent additional
  information relating these terms to each other, we can leverage this
  structural information by optimizing each term
  separately. Otherwise, as illustrated by our example, the
  optimization of one term can hamper the optimization of the other
  terms. Such functions have a block diagonal Hessian.  Conversely,
  all functions whose Hessian is everywhere block diagonal can be
  written as such separated sums
  (Appendix~\ref{app:separation}). Therefore, using a block-diagonal
  approximation of a curvature matrix is very similar to separately
  optimizing each block of variables.
\end{enumerate}

\subsection{Stepsizes for natural gradient}
\label{sec:secng}

The classic derivation of the natural gradient algorithm provides a
useful insight on the meaning of the stepsizes in gradient learning
techniques \citep{amari-nagaoka-2000,ollivier-2013}. Consider the objective
function $C(\w)=\E{E_\xi(\w)}$, where the expectation is taken over
the distribution of the examples $\xi$, and assume that the parameter space
is equipped with a (Riemannian) metric in which the squared distance between
two neighboring points $\w$ and $\w+\delta\w$ can be written as
\[
    D(\w,\w+\delta\w)^2 = \delta\w^\top G(\w)\,\delta\w + o(\|\delta\w\|^2)~.
\]
We assume that the positive symmetric matrix $G(\w)$ carries useful
information about the curvature of our objective
function,\footnote{\relax This is why this document often refer to the
  Riemannian metric tensor $G(\w)$ as the curvature matrix. This
  convenient terminology should not be confused with the notion of
  curvature of a Riemannian space.}  essentially by telling us how far
we can trust the gradient of the objective function. This leads to
iterations of the form
\begin{equation}
  \label{eq:ngproblem}
    \w\of{t+1} = \w\of{t} + \argmin_{\delta\w} \left\{
    \delta\w^\top\!\left< \nabla E(\w\of{t}) \right>
    \text{~~subject\,to~~} \delta\w^\top G(\w\of{t})\,\delta\w\leq \eta^2 \right\}~,
\end{equation}
where the angle brackets denote an average over a batch of examples and
where $\eta$ represents how far we trust the gradient in the Riemannian metric.
The classic derivation of the natural gradient reformulates this problem
using by introducing a Lagrange coefficient $1/2\gamma>0$,
\[
    \w\of{t+1} = \w\of{t} + \argmin_{\delta\w} \left\{
       \delta\w^\top\!\left<\nabla E(\w\of{t})\right>
     + \frac{1}{2\gamma} \delta\w^\top G(\w\of{t})\,\delta\w \right\}~.
\]
Solving for $\delta\w$ then yields the natural gradient algorithm
\begin{equation}
  \label{eq:ng}
    \w\of{t+1} = \w\of{t} + \gamma\, G^{-1}(\w\of{t}) \left<\nabla E(\w\of{t})\right>~.
\end{equation}
It is often argued that choosing a stepsize $\gamma$ is as good as
choosing a trust region size $\eta$ because every value of $\eta$ can
be recovered using a suitable $\gamma$. However the exact relation
between $\gamma$ and $\eta$ depends on the cost function in nontrivial
ways. The exact relation, recovered by solving
${\delta\w^\top}G(\w\of{t})^{-1} \delta\w=\eta^2$, leads to an
expression of the natural gradient algorithm that depends on $\eta$
instead of $\gamma$.
\begin{equation}
  \label{eq:ngnew}
  \w\of{t+1} = \w\of{t} + \eta \frac{ G^{-1}(\w\of{t}) \left<\nabla E(\w\of{t})\right>}
                 {\sqrt{\left<\nabla E(\w\of{t})\right>^\top\!G^{-1}(\w\of{t}) \left<\nabla E(\w\of{t})\right>}} ~.
\end{equation}
Expression \eqref{eq:ngnew} updates the weights along the same
direction as \eqref{eq:ng} but introduces an additional scalar coefficient
that effectively modulates the stepsize in a manner consistent with
Section~\ref{sec:motiv}.a. A similar approach was in advocated by
\citet{schulman-2015} for the TRPO algorithm used in Reinforcement
Learning. The next subsection shows how this approach changes when one
considers a block-diagonal curvature matrix in a manner
consistent with Section~\ref{sec:motiv}.b.

\subsection{Stepsizes for block diagonal natural gradient}
\label{sec:secngd}

We now assume that $G(\w)$ is block-diagonal. Let $\w_j$ represent
the subset of weights associated with each diagonal block $G_{jj}(\w)$.
Following Section~\ref{sec:motiv}.b, we decouple the optimization
of the variables associated with each block by replacing the natural gradient
problem~\eqref{eq:ngproblem} by the separate problems
\[
   \forall j \qquad
      \w_j\of{t+1} = \w_j\of{t} + \argmin_{\delta\w_j} \left\{
    \delta\w_j^\top\!\left<\nabla_j E(\w_t)\right>
    \text{~~subject\,to~~} \delta\w_j^\top\!G_{jj}(\w\of{t})\,\delta\w_j\leq \eta^2 \right\}~,
\]
where $\nabla_j$ represents the gradient with respect to $w_j$.
Solving as above leads to
\begin{equation}
  \label{eq:ngd}
  \forall j \qquad
  \w_j\of{t+1} = \w_j\of{t} + \eta \frac{ G_{jj}^{-1}(\w\of{t}) \left<\nabla_j E(\w\of{t})\right>}
                 {\sqrt{\left<\nabla_j E(\w\of{t})\right>^\top\!G_{jj}^{-1}(\w\of{t}) \left<\nabla_j E(\w\of{t})\right>}} ~.  
\end{equation}
This expression is in fact very similar to \eqref{eq:ngnew} except
that the denominator is now computed separately within each block,
changing both the length and the direction of the weight update.

It is desirable in practice to ensure that the denominator of
expression \eqref{eq:ngnew} or \eqref{eq:ngd} remains bounded away from zero. This is
particularly a problem when this term is subject to statistical
fluctuations induced by the choice of the batch of examples. This
can be addressed using the relation
\begin{eqnarray*}
  \left<\nabla_j E(\w\of{t})\right>^\top\!G_{jj}^{-1}(\w\of{t}) \left<\nabla_j E(\w\of{t})\right>
  &\approx& \E{\nabla_j E(\w\of{t})}^\top\!G_{jj}^{-1}(\w\of{t})\E{\nabla_j E(\w\of{t})} \\
  &\leq& \E{\nabla_j E(\w\of{t})^\top\!G_{jj}^{-1}(\w\of{t}) \nabla_j E(\w\of{t})}~.
\end{eqnarray*}
Further adding a small regularization parameter $\mu>0$ leads to the alternative formulation
\begin{equation}
  \label{eq:ngalt}
  \forall j \qquad
  \w_j\of{t+1} = \w_j\of{t} + \eta \frac{ G_{jj}^{-1}(\w\of{t}) \left<\nabla_j E(\w\of{t})\right>}
                 {\sqrt{\mu+\E{\nabla_j E(\w\of{t})^\top\!G_{jj}^{-1}(\w\of{t}) \nabla_j E(\w\of{t})}}} ~.
\end{equation}

\subsection{Recovering RMSprop}

Let us first illustrate this idea by considering the Euclidian metric~$G=I$.
Evaluating the denominator of \eqref{eq:ngalt} separately for each weight
and estimating the expectation $\E{(\nabla_jE)^2}$ with a running average
\[
   R_j\of{t} = (1-\lambda)R_j\of{t-1}+\lambda\left(\frac{\partial E}{\partial w_j}\right)^2~,
\]
yields the well-loved RMSProp weight update \citep{tieleman-hinton-2012}:
\[
   w_j\of{t+1} = w_j\of{t} -
     \frac{\eta}{\sqrt{\mu+R_j\of{t}}} \left< \frac{\partial E}{\partial w_j} \right>~.
\]

\subsection{Recovering a well-known neural network trick}
\label{sec:nntrick}

We now consider a neural network using the hyperbolic tangent
activation functions as was fashionable in the 1990s \citep{lecun-1998}.
Using the notations of Section~\ref{sec:zerooverhead},
we consider block-diagonal curvature matrices whose blocks~$G_{jj}$
are associated to the weights $\w_j=(w_{0j}\dots w_{nj})$ of each unit~$j$.
Because this activation function is centered and bounded,
it is almost reasonable to assume that the~$x_i$ have zero mean
and unit variance. Proceeding with the approximations discussed
in Appendix~\ref{app:curvature}, and further assuming the $x_i$
are uncorrelated,
\[
\big[G_{jj}\big]_{i\ip} ~\approx~ \E{g_j^2}\E{x_i x_\ip\rule{0pt}{2.2ex}}
~\approx~ \left\{\begin{array}{cl} \E{g_j^2} & \text {if $i=\ip$} \\
                  0 & \text {otherwise.} \end{array} \right.
\]
We can then evaluate the denominator of \eqref{eq:ngalt}, with $\mu=0$,
under the same approximations:
\[
  \sqrt{\frac{\E{\sum_{i=1}^n x_i g_j g_j x_i}}{\E{g_j^2}}}
  ~\approx~ \sqrt{\frac{\sum_{i=1}^n \E{x_i^2} \E{g_j^2}}{\E{g_j^2}}}
  ~=~ \sqrt{n}~.
\]
Although dividing the learning rate by the inverse square root of the
number $n$ of incoming connections (the fanin) is a well known trick for such
networks~\citep[][\S4.7]{lecun-1998}, no previous explanation
had linked it to curvature issues.

\begin{figure}[t]
  \centering
  \begin{minipage}{.49\linewidth}
    \hspace{-2em}
    \includegraphics[width=\linewidth]{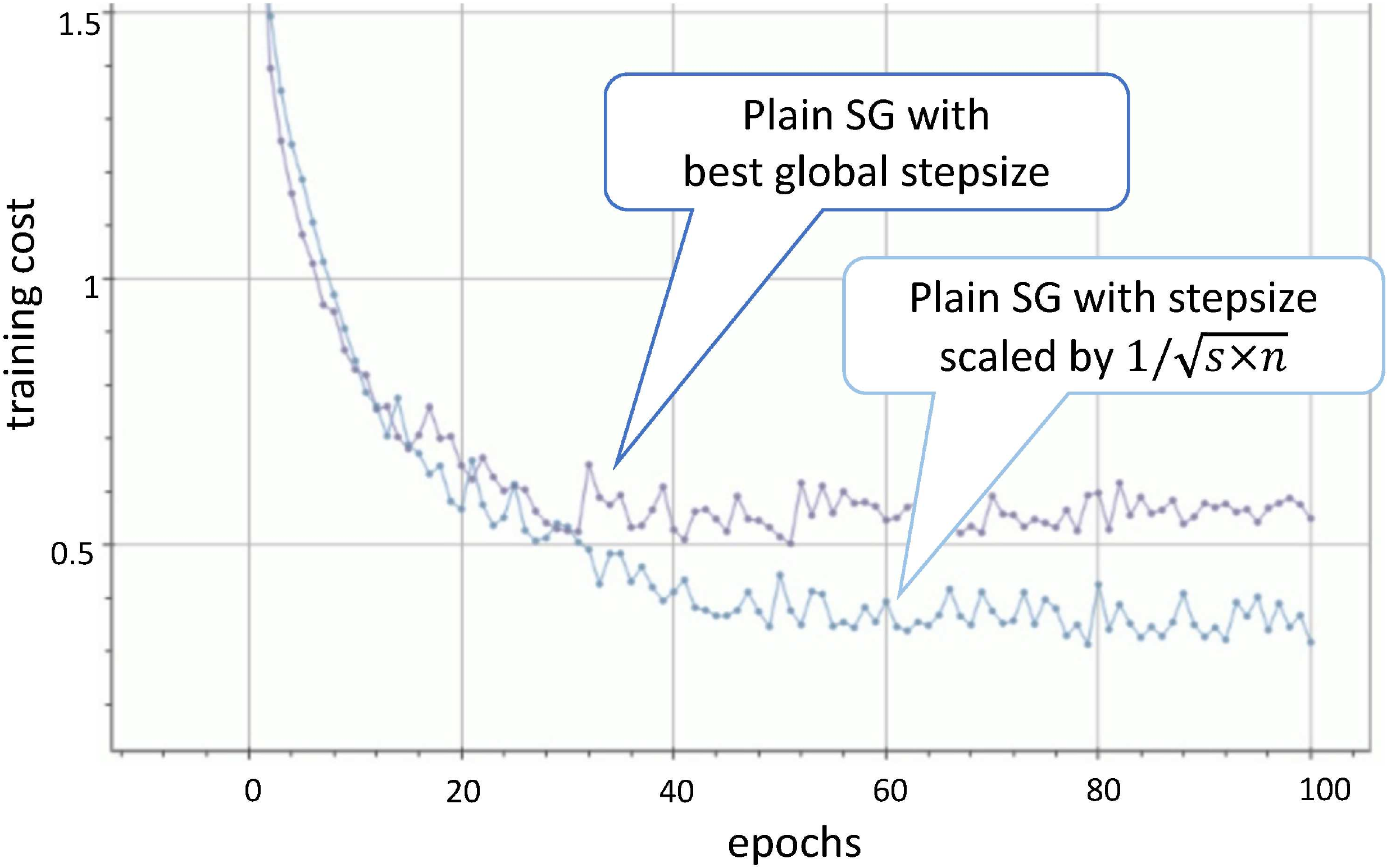}
  \end{minipage}
  \begin{minipage}{.49\linewidth}
    \includegraphics[width=\linewidth]{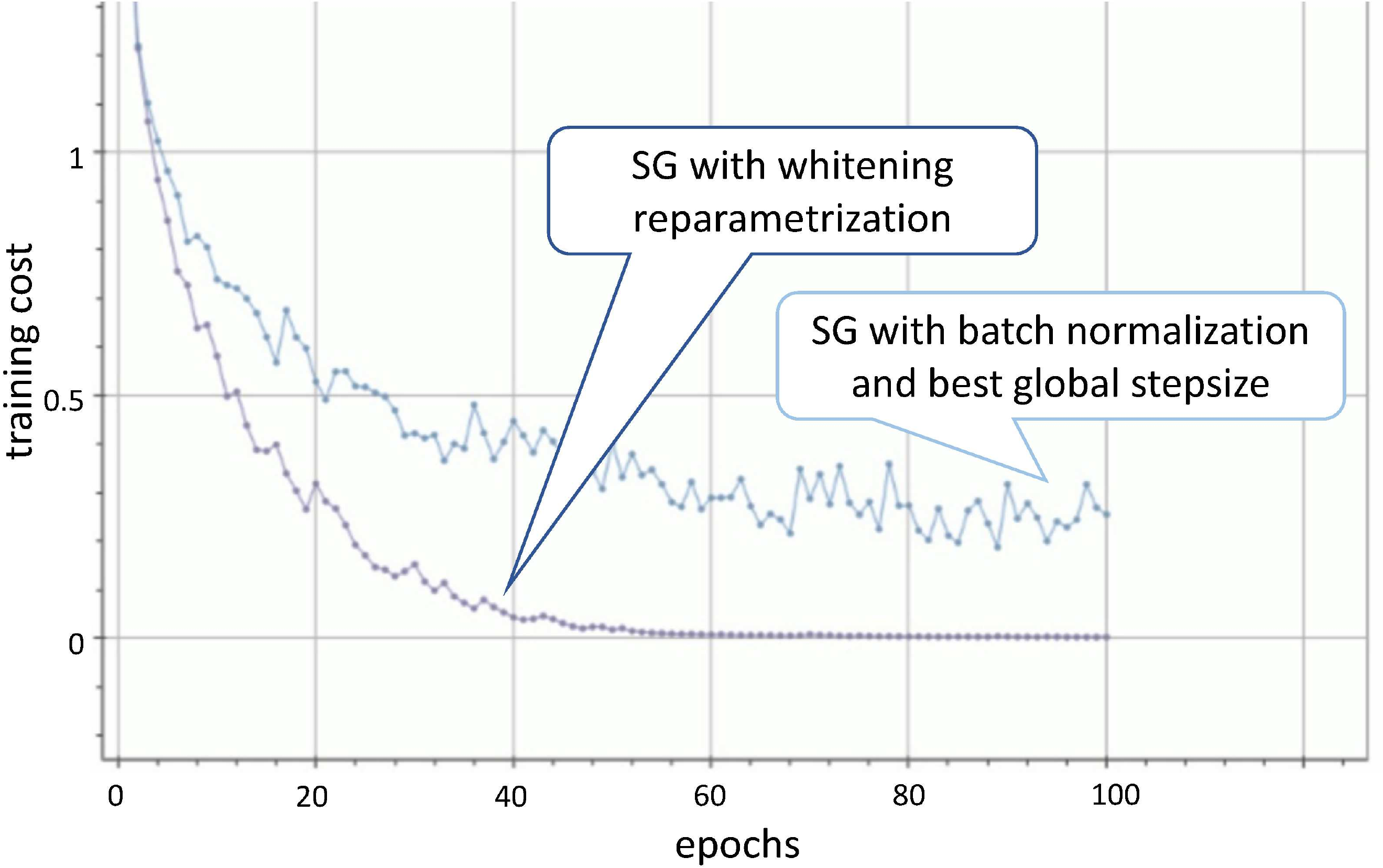}
  \end{minipage}
  \par\vspace{-1ex}
  \caption{Training a typical convolutional network (\textsf{\small C6(5x5)-P(2x2)-C16(5x5)-P(2x2)-F120-F84-F10})
    on the CIFAR10 dataset (60000 $32\times32$ color images, 10 classes).
    Left: Stochastic gradient with global stepsize and with stepsize
    divided by $\sqrt{n\times\mathcal{\scriptstyle S}}$.~ Right: Stochastic gradient
    with batch normalization versus whitening reparametrization. Note that the vertical scales
    are different.}
  \label{fig:cifar}
  \par\vspace{-1ex}
\end{figure}

Figure~\ref{fig:cifar} (left) illustrates the effectiveness of this
trick when training a typical convolutional network\footnote{
  {\small\texttt{https://github.com/soumith/cvpr2015/blob/master/Deep
      Learning with Torch.ipynb}}} on the CIFAR10 dataset. Although
our network uses ReLU instead of hyperbolic tangent activations, the
experiment shows the value of dividing the learning rates by
$\sqrt{n\times\mathcal{\scriptstyle S}}$, where $n$ represents the
fanin and where the weight sharing count $\mathcal{\scriptstyle S}$ is
always $1$ for a linear layer and can be larger for a convolutional
layer (see Appendix~\ref{app:convol}).  In both cases we use
mini-batches of 64 examples and select the global constant stepsize
that yields the best training loss after 40 epochs.
  
\section{Whitening reparametrization}
\label{sec:whitening}

Since the zero-overhead reparametrization of
Section~\ref{sec:zerooverhead} amounts to using a particular
block-diagonal curvature matrix, we can apply the insight of the
previous section and optimize the natural gradient problem within each
block. Proceeding as in Section~\ref{sec:nntrick}, we use the
reparametrization constants
\begin{equation}
  \label{eq:whiteningconstants}
  \mu_i = \E{x_i} \qquad
  \alpha^2_i=\frac{1}{\V{x_i}} \qquad
  \beta^2_j=\frac{1}{\sqrt{n\times\mathcal{\scriptstyle S}}}~,
\end{equation}
The only change relative to \eqref{eq:canonicalconstants} consists in
replacing the original $\beta^2_j=1/\E{g_j^2}$ by an expression that
depends only on the geometry of the layer (the fanin $n$ and the
sharing count $\mathcal{\scriptstyle S}$). Meanwhile the constants
$\alpha^2_i$ and $\mu_i$ are recomputed after each minibatch on the
basis of online estimates of the input moments as explained in
Section~\ref{sec:secondorder}.

An attentive reader may note that we should have multiplied (instead
of replaced) the original $\beta^2_j$ by the scaling
factor~\smash{$1/\sqrt{n\times\mathcal{\scriptstyle S}}$}. In
practice, removing the~\smash{$\E{g_j^2}$} term from the denominator
makes the algorithm more robust, allowing us to use significantly
larger global stepsizes without experiencing the occasionnal
divergences that plagued our original algorithm (the cause of this
behavior will become clearer in Section~\ref{sec:fastcurvature}.)

\subsection{Comparison with batch normalization}

Batch normalization \citep{ioffe-szegedy-2015} is an obvious point of
comparison for our reparametrization approach. Both methods attempt to
normalize the distribution of certain intermediate results. However
they do it in a substantially different way.  The whitening
reparametrization normalizes on the basis of statistics accumulated
over time, whereas batch normalization uses instantaneous statistics
observed on the current mini-batch. The whitening reparametrization
does not change the forward computation. Under batch normalization,
the output computed for any single example is affected by the other
examples of the same mini-batch. Assuming that these examples are
picked randomly, this amounts to adding a nontrivial noise to the
computation, which can be both viewed as a nuisance and as a useful
regularization technique.

\subsection{Cifar10 experiments}

Figure~\ref{fig:cifar} (right plot) compares the evolution of the
training loss of our CIFAR10 CNN using the whitening reparametrization
or using batch normalization on all layers except the output layer.
Whereas batch normalization shows a slight improvement over the
unnormalized curves of the left plot, training with the whitening
reparametrization quickly drives the training loss to zero.

From the optimization point of view, driving the training loss to zero
is a success. From the machine learning point of view, this means that
we overfit and must compensate by either adding explicit
regularization or reducing the size of the network. As a sanity check,
we have verified that we can recover the batch normalization testing
error by adding L2 regularization to the network trained with the
whitening reparametrization. The two algorithms then reduce the test error
with similar rates.\footnote{Using smaller networks would
  of course yield better speedups. A better optimization algorithm can
  conceivably help reduce our reliance on vastly overparametrized
  neural networks \citep{zhang-bengio-2017}.}

\subsection{ImageNet experiments}

In order to appreciate how the whitening reparametrization works at
scale, we replicate the above comparison using the well known AlexNet
convolutional network \citep{krizhevsky-2012} trained on ImageNet (one
million $224\times224$ training images, 1000 classes.)

The result is both disappointing and surprising. Training using only
$100,000$ randomly selected examples in ImageNet reliably yields
training curves similar to those reported in Figure~\ref{fig:cifar}
(right). However, when training on the full 1M examples, the whitening
reparametrization approach performs very badly, not even reaching the
best training loss achieved with plain stochastic gradient
descent. The network appears to be stuck in a bad place.

\subsection{Fast changing curvature}
\label{sec:fastcurvature}

The ImageNet result reported above is surprising because the
theoretical performance of stochastic gradient algorithm does not
usually depend on the size of the pool of training examples.
Therefore we spend a considerable time manually investigating this
phenomenon.

The key insight was achieved by systematically comparing the actual
statistics $\E{x_i}$ and $\V{x_i}$, estimated on a separate batch of
examples, with those estimated with the slow running average method
described in Section~\ref{sec:secondorder}. Both estimation methods
usually give very consistent results. However, in rare instance, they
can be completely different. When this happens, the reparametrization
constants $\alpha^2_i$ and $\mu_i$ are off. This often leads to
unreasonably large changes of the affected weights.  When the bias of
a particular unit becomes too negative, the ReLU activation function
remains zero regardless of the input example, and no gradient signal
can correct this in the future. In other words, these rare events
progressively disable a significant fraction of the neural network
units.

How can our slow estimation of the curvature be occasionally so wrong?
The only possible explanation is that the curvature can occasionally
change very quickly. How can the curvature change so quickly? With
a homogenous activation function like the ReLU, one does not change
the neural network output if we pick one unit, multiply its incoming
weights by an arbitrary constant $\kappa$ and divide its outgoing
weights by the same constant. This means that the cost function in
weight space is invariant along complex manifolds whose
two-dimensional slices look like hyperbolas. Although the gradient of the
objective function is theoretically orthogonal to these manifolds, a
little bit of numerical noise is sufficient to cause a movement along
the manifold when the stepsize is relatively large.\footnote{In fact
  such movements are amplified by second-order algorithms because the
  cost function has zero curvature in directions tangent to these
  manifolds.  This is why we experienced so many problems with the
  $\beta^2_j=1/\E{g_j^2}$ scaling suggested by the na\"ive second-order
  viewpoint.}  Changing the relative sizes of the incoming and
outgoing weights of a particular unit can of course dramatically
change the statistics of the unit activation.

This observation is important because most second-order optimization
algorithms assume that the curvature changes slowly
\citep{becker-lecun-1989,martens-grosse-2015,nocedal-wright-2006}.
Batch normalization does not suffer from this problem because it
relies on fresh mean and variance estimates computed on the current
mini-batch. As mentioned in Section~\ref{sec:secondorder} and
detailled in Appendix~\ref{app:coupling}, computing $\alpha_i$ and
$\mu_i$ on the current minibatch creates a nefarious feedback loop in
the training process. Appendix~\ref{app:mitigation} describes
an inelegant but effective way to mitigate this problem.

\section{Conclusion}

Investigating the robustness issues of a second-order block-diagonal
neural network stochastic gradient training algorithm has revealed two
interesting insights. The first insight reinterprets what is meant
when one makes a block-diagonal approximation of the curvature
matrix. This leads to a new way to scale the stepsizes and clarifies
popular algorithms such as RMSProp as well as old neural network
tricks such as fanin stepsize scaling. The second insight stresses the
practical importance of dealing with fast changes of the curvature.
This observation challenges the design of most second order
optimization algorithms. Since much remains to be achieved to turn
these insights into a solid theoretical framework, we believe useful
to share both the path and the insights.

\paragraph{Acknowledgments}
Many thanks to Yann Dauphin, Yann Ollivier, Yuandong Tian, and Mark Tygert for their constructive comments.

\clearpage
\bibliographystyle{plainnat}
\bibliography{norm}

\begin{thebibliography}{19}
\providecommand{\natexlab}[1]{#1}
\providecommand{\url}[1]{\texttt{#1}}
\expandafter\ifx\csname urlstyle\endcsname\relax
  \providecommand{\doi}[1]{doi: #1}\else
  \providecommand{\doi}{doi: \begingroup \urlstyle{rm}\Url}\fi

\bibitem[Amari and Nagaoka(2000)]{amari-nagaoka-2000}
Sun{-}Ichi Amari and Hiroshi Nagaoka.
\newblock \emph{Methods of Information Geometry}.
\newblock Oxford University Press, Oxford, 2000.

\bibitem[Becker and LeCun(1989)]{becker-lecun-1989}
S.~Becker and Y.~LeCun.
\newblock Improving the convergence of back-propagation learning with
  second-order methods.
\newblock In D.~Touretzky, G.~Hinton, and T.~Sejnowski, editors, \emph{Proc. of
  the 1988 Connectionist Models Summer School}, pages 29--37, San Mateo, 1989.
  Morgan Kaufman.

\bibitem[{Bottou} et~al.(2016){Bottou}, {Curtis}, and {Nocedal}]{bottou-2016}
L.~{Bottou}, F.~E. {Curtis}, and J.~{Nocedal}.
\newblock Optimization methods for large-scale machine learning.
\newblock \emph{ArXiv e-prints}, June 2016.

\bibitem[Dahl et~al.(2013)Dahl, Sainath, and Hinton]{dahl-2013}
George~E. Dahl, Tara~N. Sainath, and Geoffrey~E. Hinton.
\newblock Improving deep neural networks for {LVCSR} using rectified linear
  units and dropout.
\newblock In \emph{{IEEE} International Conference on Acoustics, Speech and
  Signal Processing, {ICASSP} 2013}, pages 8609--8613, 2013.

\bibitem[Desjardins et~al.(2015)Desjardins, Simonyan, Pascanu, and
  Kavukcuoglu]{desjardins-2015}
Guillaume Desjardins, Karen Simonyan, Razvan Pascanu, and Koray Kavukcuoglu.
\newblock Natural neural networks.
\newblock In \emph{Advances in Neural Information Processing Systems 28}, pages
  2071--2079, 2015.

\bibitem[Ioffe and Szegedy(2015)]{ioffe-szegedy-2015}
Sergey Ioffe and Christian Szegedy.
\newblock Batch normalization: Accelerating deep network training by reducing
  internal covariate shift.
\newblock In \emph{Proceedings of the 32nd International Conference on Machine
  Learning}, pages 448--456, 2015.

\bibitem[Krizhevsky et~al.(2012)Krizhevsky, Sutskever, and
  Hinton]{krizhevsky-2012}
A.~Krizhevsky, I.~Sutskever, and G.~E. Hinton.
\newblock {ImageNet Classification with Deep Convolutional Neural Networks}.
\newblock In \emph{{Advances in Neural Information Processing Systems 25}},
  2012.

\bibitem[Le~Cun et~al.(1998)Le~Cun, Bottou, Orr, and M\"uller]{lecun-etal-1998}
Y.~Le~Cun, L.~Bottou, G.~B. Orr, and K.-R. M\"uller.
\newblock Efficient backprop.
\newblock In \emph{Neural Networks, Tricks of the Trade}, Lecture Notes in
  Computer Science LNCS~1524. Springer Verlag, 1998.

\bibitem[{Le Cun} et~al.(1998){Le Cun}, Bottou, Orr, and
  M{\"{u}}ller]{lecun-1998}
Yann {Le Cun}, L\'{e}on Bottou, Genevieve~B. Orr, and Klaus-Robert
  M{\"{u}}ller.
\newblock Efficient backprop.
\newblock In \emph{Neural Networks, Tricks of the Trade}, Lecture Notes in
  Computer Science LNCS~1524. Springer Verlag, 1998.

\bibitem[Marceau{-}Caron and Ollivier(2016)]{caron-ollivier-2016}
Ga{\'{e}}tan Marceau{-}Caron and Yann Ollivier.
\newblock Practical {Riemannian} neural networks.
\newblock \emph{ArXiV CoRR}, abs/1602.08007, 2016.
\newblock URL \url{http://arxiv.org/abs/1602.08007}.

\bibitem[Martens(2014)]{martens-2014}
James Martens.
\newblock New insights and perspectives on the natural gradient method.
\newblock \emph{ArXiV CoRR}, abs/1412.1193, 2014.
\newblock URL \url{http://arxiv.org/abs/1412.1193}.

\bibitem[Martens and Grosse(2015)]{martens-grosse-2015}
James Martens and Roger~B. Grosse.
\newblock Optimizing neural networks with {Kronecker}-factored approximate
  curvature.
\newblock In \emph{Proceedings of the 32nd International Conference on Machine
  Learning (ICML 2015)}, pages 2408--2417, 2015.

\bibitem[Nocedal and Wright(2006)]{nocedal-wright-2006}
Jorge Nocedal and Stephen~J. Wright.
\newblock \emph{{Numerical {O}ptimization}}.
\newblock Springer New York, {Second} edition, 2006.

\bibitem[Ollivier(2013)]{ollivier-2013}
Yann Ollivier.
\newblock Riemannian metrics for neural networks.
\newblock \emph{ArXiV CoRR}, abs/1303.0818, 2013.
\newblock URL \url{http://arxiv.org/abs/1303.0818}.

\bibitem[Park et~al.(2000)Park, Amari, and Fukumizu]{park-amari-2000}
Hyeyoung Park, Sun{-}ichi Amari, and Kenji Fukumizu.
\newblock Adaptive natural gradient learning algorithms for various stochastic
  models.
\newblock \emph{Neural Networks}, 13\penalty0 (7):\penalty0 755--764, 2000.

\bibitem[Schraudolph(2012)]{schraudolph-2012}
Nicol~N. Schraudolph.
\newblock Centering neural network gradient factors.
\newblock In Gr{\'{e}}goire Montavon, Genevieve~B. Orr, and Klaus{-}Robert
  M{\"{u}}ller, editors, \emph{Neural Networks: Tricks of the Trade - Second
  Edition}, volume 7700 of \emph{Lecture Notes in Computer Science}, pages
  205--223. Springer, 2012.

\bibitem[Schulman et~al.(2015)Schulman, Levine, Abbeel, Jordan, and
  Moritz]{schulman-2015}
John Schulman, Sergey Levine, Pieter Abbeel, Michael~I. Jordan, and Philipp
  Moritz.
\newblock Trust region policy optimization.
\newblock In \emph{Proceedings of the 32nd International Conference on Machine
  Learning, {ICML} 2015}, pages 1889--1897, 2015.

\bibitem[Tieleman and Hinton(2012)]{tieleman-hinton-2012}
Tijmen Tieleman and Geoffrey Hinton.
\newblock Lecture 6.5. {RMSPROP}: Divide the gradient by a running average of
  its recent magnitude.
\newblock COURSERA: Neural Networks for Machine Learning, 2012.

\bibitem[Zhang et~al.(2017)Zhang, Bengio, Hardt, Recht, and
  Vinyals]{zhang-bengio-2017}
Chiyuan Zhang, Samy Bengio, Moritz Hardt, Benjamin Recht, and Oriol Vinyals.
\newblock Understanding deep learning requires rethinking generalization.
\newblock In \emph{International Conference on Representation Learning {(ICLR
  2017)}}, 2017.
\newblock Also arXiV CoRR abs/1611.03530.

\end{thebibliography}

\clearpage
\begin{appendices}

\section{Derivation of the curvature matrix}
\label{app:curvature}

For the sake of simplicity, we only take into account the parameters
$\v=(\dots v_\ij \dots)$ associated with a particular linear layer of
the network (hence neglecting all cross-layer interactions).  Each
example is then represented by the layer inputs~$x_i$ and by an
additional variable $\xi$ that encode any relevant information not
described by the~$x_i$.  For instance $\xi$ could represent a class
label.  We then assume that the cost function associated with a single
example has the form
\[
   E(\v; \xi, x_1\dots x_n)
     = -\log\big( \varphi(\xi, y_1\dots y_m) \big)
     = -\log\left( \varphi\left(\xi, \dots \beta_j \sum_{i=0}^n v_\ij x_i \dots \right) \right),
\]
where the function $\varphi$ encapsulates all the layers following the
layer of interest as well as the loss function.  This kind of cost
function is very common when the quantity $\varphi$ can be interpreted
as the probability of some event of interest.

The optimization objective $C(\v)$ is then the expectation of $E$ with
respect to the variables $\xi$ and $x_i$,
\[
   C(\v) = \E{E(\xi,x_1\dots x_n)} .
\]
Its derivatives are
\[
   \frac{\partial C}{\partial v_\ij}
   ~=~ \E{ - \frac{1}{\varphi} \frac{\partial \varphi}{\partial y_j} \beta_j z_i }
   ~=~ \E{ g_j \beta_j z_i }
\]   
and the coefficients of its Hessian matrix are
\[
   \frac{\partial^2 C}{\partial v_\ij \partial v_\ipjp}
   ~=~ \E{ \left( \frac{1}{\varphi^2}
                \frac{\partial \varphi}{\partial y_j}
                \frac{\partial \varphi}{\partial y_\jp}
                - 
               \frac{1}{\varphi}
               \frac{\partial^2 \varphi}{\partial y_j y_\jp}
         \right) \: \beta_j \beta_\jp z_i z_\ip
       } ~.
   \]
   
Our first approximation consists in neglecting all the terms of the
Hessian involving the second derivatives of $\varphi$, leading
to the so-called \emph{Generalized Gauss-Newton} matrix $G$
\cite[\S 6.2]{bottou-2016}
whose blocks $G_{j\jp}$ have coefficients
\[
    \big[G_{j\jp}\big]_{i\ip} ~=~ \E{ \frac{1}{\varphi^2}
         \frac{\partial \varphi}{\partial y_j}
         \frac{\partial \varphi}{\partial y_\jp}
          \beta_j \beta_\jp z_i z_\ip
   } ~=~ \beta_j \beta_\jp \E{ g_j g_\jp z_i z_\ip}
\]
Interestingly, this matrix is exactly equal to a well known
approximation of the Fisher information matrix called the
\emph{Empirical Fisher} matrix \citep{park-amari-2000,martens-2014}.
    
We then neglect the non-diagonal blocks and assume that
the squared gradients $g_j^2$ are not correlated with either
the layer inputs $x_i$ or their cross products $x_ix_\ip$.
See \citep{desjardins-2015} for a similar approximation.
Recalling that $z_0=1$ is not correlated with anything by definition,
this means that the $g_j^2$ is not correlated with $z_iz_\ip$ either.
\[
    \big[G_{jj}\big]_{i\ip} ~=~ \beta_j^2 \E{g_j^2} \E{z_iz_\ip}~.
\]
Further assuming that the layer inputs $x_i$ are also decorrelated
leads to our final expression
\[
   \big[G_{jj}\big]_{i\ip} ~=~ \beta_j^2 \E{g_j^2} \times \left\{ \begin{array}{ll}
     \E{z_i^2} & \text{if $i=\ip$,} \\[1ex]
     \E{z_i}\E{z_\ip} & \text{otherwise.}
   \end{array}\right.
\]

The validity of all these approximations is of course questionable.
Their true purpose is simply to make sure that our approximate
curvature matrix $G$ can be made equal to the identity with a simple
choice of the reparametrization constants, namely,
\[
   \mu_i=\E{x_i} \qquad
   \alpha^2_i=\frac{1}{\V{x_i}} \qquad
   \beta^2_j=\frac{1}{\E{g_j^2}} ~.
\]

\section{Reparametrization of convolutional layers}
\label{app:convol}
  
Convolutional layers can be reparametrized in the same manner as
linear layers (Section~\ref{sec:zerooverhead}) by introducing
additional indices $u$ and $v$ to represent the two dimensions of the
image and kernel coordinates. Equations \eqref{eq.linear} and
\eqref{eq:reparam} then become
\begin{eqnarray*}
  y_{ju_1u_2}
  &=& w_{0j}+\sum_{i=1}^{n} ~ \sum_{v_1v_2}
          x_{i(u_1+v_1)(u_2+v_2)}\:w_{ijv_1v_2} \\
  &=& \beta_j\left( v_{0j} +  \sum_{i=1}^{n} ~ \sum_{v_1v_2}
          \alpha_i\big(x_{i(u_1+v_1)(u_2+v_2)}-\mu_i\big)\:v_{ijv_1v_2}\:\right) ~,
\end{eqnarray*}
and the derivative of the loss $E$ with respect to a particular weight 
involves a summation over all the terms involving that weight:
\[
 \frac{\partial E}{\partial v_{i j v_1 v_2}} 
  =  \beta_j \: \sum_{u_1u_2}  g_{ju_1u_2}\:z_{i(u_1+v_1)(u_2+v_2)}  ~.
\]
Following Appendix~\ref{app:curvature}, we write the blocks $G_{jj'}$ of the generalized Gauss Newton matrix $G$,
\[
   \left[G_{jj'}\right]_{iv_1v_2,i'v_1'v_2'} = \E{ \frac{\partial E}{\partial v_{i j v_1 v_2}} \frac{\partial E}{\partial v_{i'j'v_1'v_2'}} }\,.
\]
Obtaining a convenient approximation of $G$ demands questionable assumptions such as neglecting 
nearly all off-diagonal terms, and nearly all possible correlations involving the $z$ and $g$ variables.  
This leads to the following choices for the reparametrization constants, where the expectations and 
variances are also taken across the image dimension subscripts (``$\bullet$'') and where the 
constant $\mathcal{\scriptstyle S}$ counts the number of times each weight is shared, that is,
the number of applications of the convolution kernel in the convolutional layer.
\[
   \mu_i=\E{x_{i\bullet\bullet}} \qquad
   \alpha^2_i=\frac{1}{\V{x_{i\bullet\bullet}}} \qquad
   \beta^2_j=\frac{1}{\mathcal{\scriptstyle S}\: \E{g_{j\bullet\bullet}^2}} ~.
\]

\section{Coordinate separation}
\label{app:separation}

It is obvious that a twice differentiable function
\[
   f~:~ (x_1\dots x_k)\in\R^{n_1}\times\dots\times\R^{n_k} ~ \longmapsto f(x_1,\dots,x_k)\in\R
\]
that can be written as a sum
\begin{equation}
  \label{eq:separated}
  f(x_1\dots x_k) = f_1(x_1)+\dots+f_k(x_k)
\end{equation}
has a block diagonal Hessian everywhere, that is,
\begin{equation}
  \label{eq:bdhessian}
  \forall (x_1\dots x_k)\in\R^{n_1}\times\dots\times\R^{n_k} \quad \forall i \neq j \quad
      \frac{\partial^2 f}{\partial x_i \partial x_j} = 0~.
\end{equation}
Conversely, assume the twice differentiable function $f$ satisfies \eqref{eq:bdhessian}, and write
\begin{eqnarray*}
  f(x_1\dots x_k) - f(0\dots 0)
  &=& \sum_{i=1}^{k} f(x_1\dots x_i,0\dots 0) - f(x_1\dots x_{i-1},0\dots 0) \\
  &=& \sum_{i=1}^{k} \int_0^1 x_i^\top\:\frac{\partial f}{\partial x_i}(x_1\dots x_{i-1}, t x_i, 0\dots 0)\:dt~.
\end{eqnarray*}
Then observe
\begin{eqnarray*}
  \lefteqn{\frac{\partial f}{\partial x_i}(x_1\dots x_{i-1}, r, 0, \dots 0)
     - \frac{\partial f}{\partial x_i}(0\dots 0,r,0\dots 0) } \\
  &=&  \int_0^1 \sum_{j=1}^{i-1} x_j^\top\:\frac{\partial^2 f}{\partial x_j\partial x_i}(tx_1\dots tx_{i-1},r,0\dots 0)\: dt
  ~=~ 0~.
\end{eqnarray*}
Therefore property \eqref{eq:separated} is true because
\[
   f(x_1\dots x_k) = f(0\dots 0) + \sum_{i=1}^{k}
     \int_0^1 x_i^\top\:\frac{\partial f}{\partial x_i}(0\dots 0, t x_i, 0, \dots 0)\:dt ~.
\]

\section{Coupling effects when adapting reparametrization constants}
\label{app:coupling}

The reparametrization constants suggested by
\eqref{eq:canonicalconstants} and \eqref{eq:whiteningconstants} are
simple statistical measurements on the network variables. It is
tempting use to directly compute estimates $\hat{\alpha}_i$,
$\hat{\mu}_i$, and $\hat{\beta}_j$ on the current mini-batch in a
manner similar to batch renormalization. 

Unfortunately these estimates often combine in ways that create
unwanted biases. Consider for instance the apparently benign case
where we only need to compute an estimate $\hat{\mu}_i$ because an
oracle reveals the exact values of $\alpha_i$ and $\beta_j$.
%
Replacing $\mu_i$ by its estimate $\hat{\mu}_i$ in the update
equations~\eqref{eq:quasidiagonal} gives the actual weight updates
$\widehat{\delta w_\ij}$ performed by the algorithm. Recalling that
$\hat{\mu}_i$ is now a random variable whose expectation is $\mu_i$,
we can compare the expectation of the actual weight
update~{$\E{\widehat{\delta w_\zj}}$} with the ideal value~$\E{\delta
  w_\zj}$.
\begin{align*}
  \E{\widehat{\delta w_\zj}}
  &= \E{\beta_j^2 g_j \left(1 - \sum_i \alpha_i^2 \hat{\mu}_i (x_i-\hat{\mu}_i)\right)} \\
  &= \beta_j^2 \left( 1 - \sum_i \alpha_i^2 \left( \E{\hat{\mu}_i x_i g_j} - \E{\hat{\mu}_i^2 g_j} \right) \right)\\
  &= \E{\delta w_\zj} + \sum_i \beta_j^2 \alpha_i^2 \left(
       \V{\hat{\mu}_i} \E{g_j} + \mathrm{cov}[\hat{\mu}_i^2, g_j] - \mathrm{cov}[\hat{\mu}_i, x_i g_j] \right)~.
\end{align*}
This derivation reveals a systematic bias that results from the
nonzero variance of $\hat{\mu_i}$ and its potential correlation with
other variables.  In practice, this bias is more than sufficient to
severely disrupt the convergence of the stochastic gradient algorithm.

\section{Mitigating fast curvature change events}
\label{app:mitigation}

\begin{figure}[h]
  \centering
  \includegraphics[width=0.5\linewidth]{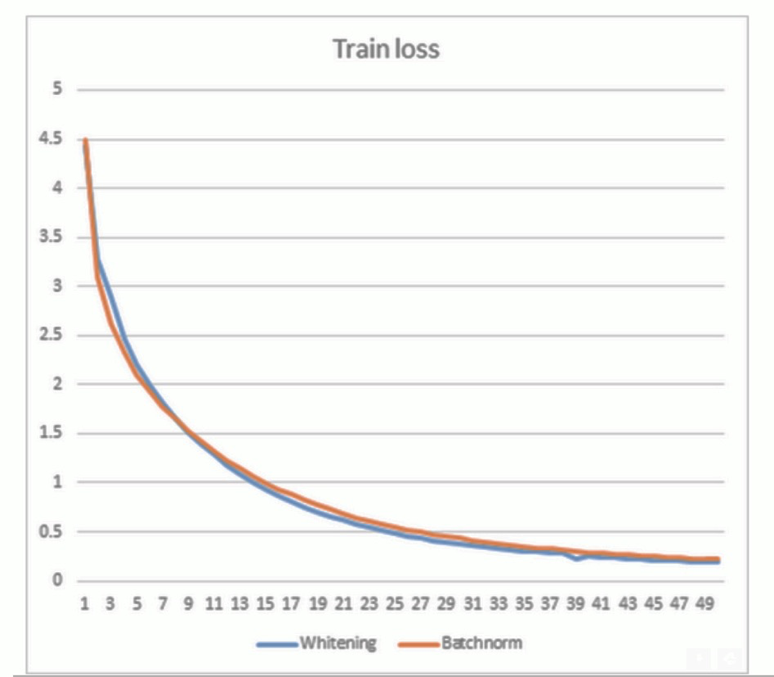}
  \caption{Mitigating fast curvature change events by using
    batch-normalization during the first epoch then either switching
    to the whitening reparametrization (blue curve) or keeping the
    batch normalization (blue curve).  Although both methods appear
    similar in terms of number of epochs, the whitening
    reparametrization implementation is faster than the optimized
    batch normalization implementation. Note that the training loss in
    this curve was estimated after each epoch by performing a full
    sweep on the training data (unlike figure~\ref{fig:cifar} which
    plots an estimate of the loss computed while training.)}
  \label{fig:mitigation}
\end{figure}

Fast curvature changes mostly happens during the first phase of the
training process and disappears when the training loss stabilizes. For
ImageNet, we were able to mitigate the phenomenon by using
batch-normalization during the first epoch then switching to the
whitening reparametrization approach for the remaining epochs
(Figure~\ref{fig:mitigation}.)

\end{appendices}

\end{document}